\PassOptionsToPackage{table,dvipsnames,xcdraw}{xcolor}
\documentclass[sigconf]{acmart}

\usepackage[utf8]{inputenc} %
\usepackage[T1]{fontenc}    %
\usepackage{hyperref}       %
\usepackage{url}            %
\usepackage{booktabs}       %
\usepackage{nicefrac}       %
\usepackage{microtype}      %

\usepackage{bbm}
\usepackage{dsfont}

\usepackage{algorithmic}
\usepackage{graphicx}

\usepackage{adjustbox}
\usepackage{float}
\usepackage{breqn} 

\usepackage{enumitem}

\usepackage{xurl}  %
\usepackage{seqsplit}  %
\usepackage{hyphenat}  %

\usepackage{tikz}
\usetikzlibrary{positioning, arrows.meta}
\usepackage{pgfplots}
\pgfplotsset{compat=1.17}

\newcommand{\btt}[1]{\texttt{\seqsplit{#1}}}

\usepackage{listings}

\definecolor{authcolor}{RGB}{220,20,60}       %
\definecolor{scopecolor}{RGB}{0,128,0}        %
\definecolor{limitcolor}{RGB}{255,140,0}      %
\definecolor{filtercolor}{RGB}{70,130,180}    %
\definecolor{attrcolor}{RGB}{138,43,226}      %
\definecolor{securitycolor}{RGB}{184,134,11}  %

\copyrightyear{2026}
\acmYear{2026}
\setcopyright{cc}
\setcctype{by-nc-nd}
\acmConference[KDD '26]{Proceedings of the 32nd ACM SIGKDD Conference on Knowledge Discovery and Data Mining V.2}{August 09--13, 2026}{Jeju Island, Republic of Korea}
\acmBooktitle{Proceedings of the 32nd ACM SIGKDD Conference on Knowledge Discovery and Data Mining V.2 (KDD '26), August 09--13, 2026, Jeju Island, Republic of Korea}
\acmDOI{10.1145/3770855.3818449}
\acmISBN{979-8-4007-2259-2/2026/08}



\title{ML-Powered LDAP Reconnaissance Detection using Weak Supervision}

\author{Shaefer Drew}
\orcid{0009-0005-6710-8408}
\affiliation{%
  \institution{CrowdStrike}
  \city{Redmond}
  \state{OR}
  \country{USA}
}
\email{shaefer.drew@crowdstrike.com}

\author{Edward Raff}
\orcid{0000-0002-9900-1972}
\affiliation{%
  \institution{CrowdStrike}
  \city{Jamesville}
  \state{NY}
  \country{USA}
}
\email{edward.raff@crowdstrike.com}

\author{Michael Brautbar}
\orcid{0009-0004-7776-4698}
\affiliation{%
  \institution{CrowdStrike}
  \city{Groton}
  \state{MA}
  \country{USA}
}
\email{michael.brautbar@crowdstrike.com}

\author{Yaron Zinar}
\orcid{0009-0000-7889-7661}
\affiliation{%
  \institution{CrowdStrike}
  \city{Petah Tikva}
  \country{Israel}
}
\email{yaron.zinar@crowdstrike.com}

\author{Benjamin Malmberg}
\orcid{0000-0002-2720-7963}
\affiliation{%
  \institution{CrowdStrike}
  \city{Utrecht}
  \country{Netherlands}
}
\email{benjamin.malmberg@crowdstrike.com}

\author{Dor Agron}
\orcid{0009-0001-8133-586X}
\affiliation{%
  \institution{CrowdStrike}
  \city{Zurich}
  \country{Switzerland}
}
\email{dor.agron@crowdstrike.com}

\author{Sagi Sheinfeld}
\orcid{0009-0006-1688-1975}
\affiliation{%
  \institution{CrowdStrike}
  \city{Tel-Aviv}
  \country{Israel}
}
\email{sagi.sheinfeld@crowdstrike.com}

\author{Avraham Kama}
\orcid{0009-0009-7357-0796}
\affiliation{%
  \institution{CrowdStrike}
  \city{Beer Yaakov}
  \country{Israel}
}
\email{avi.kama@crowdstrike.com}

\author{Asaf Romano}
\orcid{0009-0001-0045-0517}
\affiliation{%
  \institution{CrowdStrike}
  \city{Giv'atayim}
  \country{Israel}
}
\email{asaf.romano@crowdstrike.com}

\renewcommand{\shortauthors}{Shaefer Drew et al.}

\begin{document}

\begin{abstract}
Lightweight Directory Access Protocol (LDAP) is a protocol that allows users to query and modify Active Directory (AD) data. By default, all users have read access to all AD data through LDAP, making it a common initial tool for reconnaissance when a threat actor first compromises an identity. To capture threat actors early in the reconnaissance phase, we developed two machine learning frameworks to detect LDAP reconnaissance: an ML classifier to predict malicious LDAP queries and an ML-based data-mining method to extract malicious query signatures. By correlating LDAP queries with endpoint detections, the first framework uses weak supervision to label a massive dataset and classify LDAP queries as malicious or benign. For immediate deployment, a second technique was developed on top of this approach to employ a rigorous statistical hypothesis-testing framework for mining novel, malicious LDAP signatures. While this weakly supervised approach is limited compared with manual human labeling, it is more practical for this use case because it leverages large-scale automated corpus construction, reducing costs and time. Ultimately, both the LDAP classifier and the ML-based LDAP signature mining method achieved performance benchmarks, with the classifier achieving up to a 65\% True Positive Rate (TPR) on the holdout set while limiting false positives, and mined signatures demonstrating 81.48\% field precision with CrowdStrike's Managed Detection and Response team.
\end{abstract}

\maketitle

\section{Introduction}

Early in the attack chain, reconnaissance is the first tactic used by threat actors to gather critical information about the target environment and plan their attack. During this phase, adversaries map networks, enumerate accounts, and identify vulnerabilities without triggering alarms. Proactive monitoring and early detection of this activity can disrupt threat actors in their tracks and stop a breach long before their malicious operations \cite{drew2025ldap}.

In this paper, we detail a new approach to detecting early signs of reconnaissance queries using AI. We specifically focus on \textit{Lightweight Directory Access Protocol (LDAP)}, which is a protocol that allows users to query and modify Active Directory (AD) data (endpoint, user, and group data). While modifications require administrative privileges, all AD users can read all AD data by default. This makes LDAP a common target for reconnaissance attacks following an initial compromise of a network.
Existing LDAP reconnaissance detections rely primarily on signature matching of known attack tools, which lack adaptive capabilities and pattern recognition for detecting lesser known attack tools, manual queries, or novel reconnaissance methods.

This paper presents 2 frameworks for detecting malicious LDAP reconnaissance. Framework 1 uses an ML classifier to predict malicious LDAP queries. Due to deployment constraints at the time of development, a 2nd framework, ML-Based LDAP Signature Mining, was created to integrate immediately with the existing architecture and protect customers while minimizing false positives.
Both frameworks leverage the same LDAP search-derived features, including text embeddings from the query and attributes, as well as numerical and categorical encodings derived from the metadata and search specifications. 

\begin{figure}[!h]
    \centering
    \includegraphics[width=\linewidth]{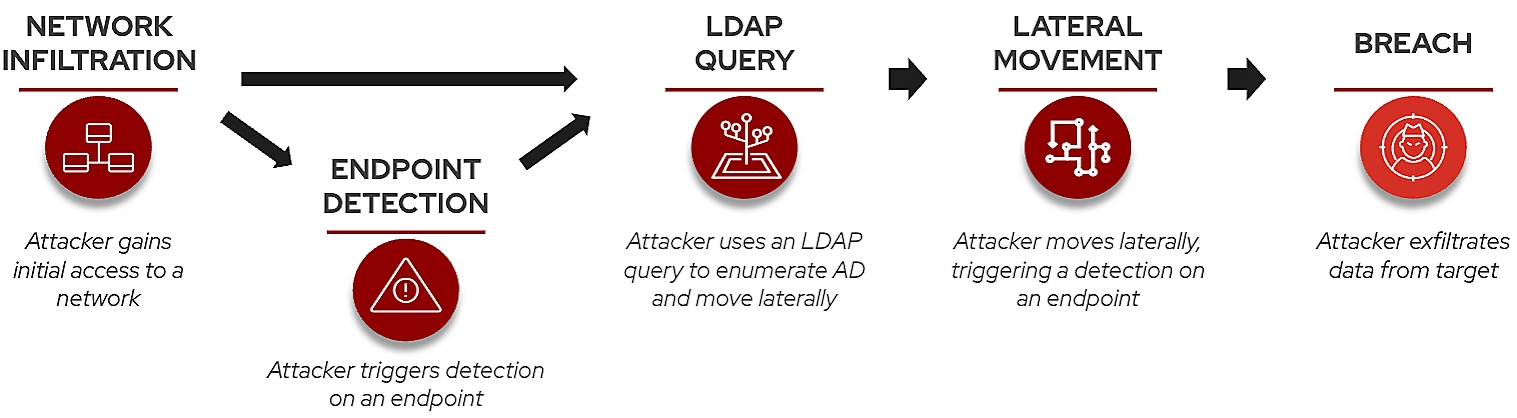}
    \caption{Identity Attack Kill Chain. This shows how threat actors use LDAP to move laterally in customer environments. They can trigger endpoint detections before and/or after LDAP enumeration, signifying the logic behind correlating endpoint detections for labeling.}
    \label{fig:kill_chain}
\end{figure}

Each framework also uses weak supervision to label and classify a large sample of LDAP queries. \textit{Weak Supervision} is a methodology where noisier sources of labels are used to assemble a much larger training corpus than manual labeling \cite{snorkel2022weak, Ratner_2017}. The weak supervised labeling employed in CrowdStrike's LDAP ML detections revolves around the attack chain shown in Figure \ref{fig:kill_chain}. This demonstrates how threat actors exploit LDAP to move laterally within customer environments. They can trigger endpoint detections before and/or after LDAP enumeration. For example, a user may gain initial access to a network and use LDAP to enumerate all domain admins. They may then try to pivot to a domain admin account by attempting to harvest credentials, triggering endpoint detection for credential theft. Since this same user queried LDAP around the same time they were alerted for credential harvesting, an inference can be made that the LDAP query itself was used for reconnaissance and can be labeled malicious. 

Framework 1 fits an ML classifier on these weak labels, using grouped splitting to generalize across different users and tuning the model according to identity standards. Section \ref{sec:framework_1} goes into detail about the methods of framework 1, including its weak labeling logic and downstream supervised classification. 

During development, the ML classifier could not be deployed to production due to architectural constraints. This revealed a coverage gap in detecting LDAP reconnaissance, leaving customers vulnerable. To protect customers immediately, a second framework was created to mine for signatures that could be deployed immediately using existing architecture. Section \ref{sec:framework_2} elaborates on this signature mining framework, detailing the training and inference of 5 downstream classifiers and signature exclusion via hypothesis testing. 

Finally, both frameworks are evaluated using multiple metrics on historical and live data. The classifier framework demonstrates effectiveness by achieving its benchmark goal of 30\% True Positive Rate (TPR) and a False Positive Rate (FPR) corresponding to less than 1 per day per customer. The signature mining focuses more on precision and low FPR, ensuring the same FPR through Family-Wise-Error-Rate (FWER) exclusion and achieving 81.48\% precision in the field. 

By correlating endpoint detections with LDAP queries, this weakly supervised machine learning method offers a robust, generalizable solution for detecting compromised identities early in the attack chain.

\section{Related Work}

Automated signature generation, essentially a kind of regular expression bespoke to a cybersecurity context, has been studied for decades primarily in the context of executable binary programs~\cite{Kim2004,Newsome:2005:PAG:1058433.1059393,Yegneswaran:2005:AGS:1251398.1251405,Blichmann2008,Venkataraman2008,li_packgenome_2023}, usually relying on search algorithms like longest common subsequence and domain knowledge parsing of the content (e.g., disassembling files and placing wild-cards in instruction arguments). Though some works have explored network~\cite{Rafique:2013:FMC:2941590.2941601,kapoor_rexactor_2021}, to wit, our work appears to be among the earliest to tackle LDAP specifically. 
The use of machine learning techniques to help generate signatures started with the use of bi-clustering to group candidate features (n-grams) together~\cite{Raff2020autoyara}. Others have since leveraged classical rule-mining~\cite{du_autocombo_2021,kapoor_rexactor_2021}, fuzzers~\cite{chen_jigsaw_2022}, and Large Language Models (LLMs)~\cite{balasubramanian_hex2sign_2024,zhang_automatically_2025}. In contrast to these approaches, we use a simpler method to pre-label our data in a weak manner and must account for operational constraints to avoid data leakage unique to LDAP queries. 

Existing signature-matching approaches to malicious LDAP queries target known attack tools. Common tools include SharpHound, Rubeus, SOAPHound, and others \cite{SharpHound, Rubeus, SOAPHound, BloodHound, BloodHound.py}. Most of these tools are open-source, facilitating monitoring and signature generation from their queries. This approach is highly precise but is limited to a small subset of attack-tool traffic (i.e., low recall). It doesn't generalize to unseen queries and can be easily evaded through slight changes in the query. Below is an example of a malicious LDAP query (1) from the Rubeus attack tool. This query enumerates SAM accounts of a certain type that do not require Kerberos pre-authentication. LDAP query 2 below performs the same operation; however, it goes undetected because its signature changes after a slight modification that specifies \textcolor{red}{(objectCategory=user)}. Both queries return users that don't require Kerberos pre-auth, but the 2nd query evades detection because it isn't part of a known attack tool.

\begin{itemize}
    \item \textbf{1: Rubeus Malicious LDAP Query}:
    \begin{lstlisting}[basicstyle=\small\ttfamily, breaklines=true]
(&(samAccountType=805306368)(userAccountControl:1.2.840.113556.1.4.803:=4194304))
    \end{lstlisting}
    
    \item \textbf{2: Slightly Changed Manual Malicious LDAP Query}:
    \begin{lstlisting}[basicstyle=\small\ttfamily, breaklines=true, escapeinside={(*}{*)}]
(&(*\textcolor{red}{(objectCategory=user)}*)(samAccountType=805306368)(userAccountControl:1.2.840.113556.1.4.803:=4194304))
    \end{lstlisting}

\end{itemize}

Since these signatures don't generalize, they can be easily evaded by such techniques. Signature mapping is limited to known attack tools and may miss manual queries, modified attack-tool traffic, and lesser-known attack methods \cite{drew2025ldap}.  

Manually adding new signatures is also time-consuming. Implementing a new signature requires a thorough investigation and review process to ensure low false-positive rates and high-precision signatures associated with attack tools. This results, on average, in only $\approx$5 new signatures annually produced through these manual methods. Furthermore, just because these signatures trigger doesn't mean they are malicious. SOC analysts must still investigate the query and the user's context to make a final determination of maliciousness. With this lengthy manual process of labeling queries, the number of ground-truth labels available to train a supervised classifier is limited.

An existing alternative to attack tool signatures is static rules for detecting LDAP reconnaissance. This approach categorizes specific types of LDAP searches (i.e., Service Principal Name (SPN) enumeration, Access Control List (ACL) enumeration, etc.). While this approach categorizes LDAP search action types with high accuracy, it demonstrates limited discriminative power between malicious and benign queries, as legitimate administrative tasks often involve identical search operations. For example, ACL enumeration can serve as a means of conducting malicious reconnaissance, but it can also be a common task for a System Administrator managing access and permissions. 

Our ML-Based frameworks address these gaps by building a more adaptable and scalable solution that can identify and respond to unfamiliar or custom-built reconnaissance techniques. These frameworks learn generalizable patterns across different signatures, preventing common evasion techniques, such as the one mentioned above, while also detecting novel reconnaissance attempts. Using weak supervision, our framework generates a large labeled corpus of millions of LDAP queries from which the model learns to distinguish malicious from benign queries.

\section{Data}

For readers who are not familiar with LDAP, we break down a malicious LDAP query and how to read it in a mini-tutorial in Appendix \ref{sec:ldap_tutorial}. The initiative started with a dataset of $\approx$6 billion LDAP queries, across 367 customers, and over the course of 90 days. This dataset then underwent filtering criteria for "uncommonness", which eliminated LDAP queries that re-occurred within a specific, recent timeframe for that customer. This predefined threshold, withheld for security concerns\footnote{We are describing a real system used in production, and our industry is inherently adversarial. So we must balance exposition with giving away important details that would allow malicious actors to more easily subvert our systems.}, was derived from Identity Protection subject matter experts. Post-filtering, this resulted in $\approx$1.3 million LDAP queries to be labeled by the heuristic rules. This large dataset enabled us to effectively mine for correlated endpoint detections and to obtain a representative sample from which the model can be trained. The heuristic rules were able to parse through this dataset and label all of the LDAP queries in a matter of minutes, something which would be impractical for human labelers.

\section{Framework 1: ML LDAP Classification}
\label{sec:framework_1}

We present an initial ML classification framework for suspicious LDAP detection. Due to deployment constraints, a second framework is built on the first to mine for suspicious LDAP query signatures. This section will present the first classification framework in full.

The LDAP detection classification framework includes 2 steps:
\begin{enumerate}
    \item \textbf{Weak Labeling:} LDAP queries are joined with endpoint detections in close proximity for the same source user in order to produce a binary label.
    
    \item \textbf{Downstream Supervised Classification:} A supervised classifier is fit on the weak labels to predict maliciousness.
\end{enumerate}

\subsection{Weak Labeling}

This framework uses \textit{weak supervision}, a methodology where noisier sources of labels are used to assemble a much larger training corpus than manual labeling \cite{snorkel2022weak, Ratner_2017}. These noisier sources often come in the form of heuristic rules.

Manual LDAP query labeling would involve humans reviewing millions of queries, analyzing their context, and labeling them individually as malicious or benign. The current method for discovering and validating new attack tool signatures produces $\approx$5 new signatures per year. Since most LDAP queries are benign and there's a clear human resource constraint for manual labeling, we found manual labeling not to be a viable alternative.  To create heuristic criteria for automatically labeling over a million LDAP queries, we turned to correlating identity data (LDAP queries) with endpoint data (Endpoint Detections). 

As shown in Figure \ref{fig:kill_chain}, LDAP is a common tool for reconnaissance that can occur both before and after endpoint detections. Using this knowledge, LDAP queries were labeled as malicious according to the following criteria:

\begin{enumerate}
    \item \textbf{Query relates to a high-confidence endpoint detection:} The LDAP query must be associated with a high-efficacy detection (one with a low false positive rate), a detection commonly seen with LDAP reconnaissance in the wild, and it must make intuitive sense as an attack path.

    Distilling high-confidence detections involved consulting threat hunters and analyzing the field efficacy of selected detections. 31 high-confidence endpoint detections were selected based on their high field precision, correlation with LDAP queries, and attack-chain feasibility. These were used to label LDAP queries. This included, but wasn't limited to, ML-based detections for malicious binaries, WMI-related detections, credential scanning detections, registry credential-related detections, LSASS detections, PowerShell detections, psexec detections, RDP detections, and SMB detections.

    \item \textbf{Query occurs within a close time frame of an endpoint detection:} The LDAP query must have occurred within a close time frame of an overlapping endpoint detection associated with the same user.

    This criterion relied on subject experts analyzing the results of their 
    investigations and determining the time frame during which most 
    LDAP-Endpoint correlations occur. Based on threat hunter investigation 
    patterns, analysts routinely inspect the hour preceding a suspicious 
    endpoint detection when investigating lateral movement. This 1-hour 
    window was therefore adopted as the labeling criterion. A tighter window 
    would yield higher-confidence labels but fewer training examples, while 
    a wider window would introduce noise from benign queries incorrectly 
    co-labeled with detections. This tradeoff informed the choice of 1 hour 
    as a principled balance between label quality and corpus size.

    \item \textbf{Query is uncommon:} The query should be relatively uncommon or anomalous for that user to be executing, based on a baseline of that customer's query activity.

    Criterion 3 helped filter out noise by examining a customer's LDAP query history and determining whether the query was out of the ordinary. This involved getting a rolling count of the \btt{LdapSearchQueryToken} in question over the past n days for the customer on which the query was executed. If the count was at or above a certain threshold (redacted for security purposes), then the query was deemed common and excluded from the dataset. If it fell below the threshold, it was deemed uncommon and was included in the training dataset. This "uncommonness" filter reduced the $\approx$6 billion initial LDAP queries to $\approx$1.3 million.
\end{enumerate}

These labeling criteria allow us to correlate endpoint detections, thereby providing historical LDAP queries with automatic labels of malicious or benign. This enables the rapid generation of a large sample set of malicious LDAP queries for training a supervised classifier.

\subsection{Downstream Supervised Classification}

\subsubsection{Grouped Splitting}

LDAP reconnaissance is considered a Living off the Land attack because threat actors exploit an existing Active Directory function for malicious purposes. LDAP is designed for authentication and access control, and for managing permissions and user information. However, threat actors can use this same retrieval functionality to enumerate a customer environment and move laterally. Sometimes an LDAP query may be considered benign because it's used for legitimate purposes, such as IT administration; however, the same query may be considered malicious when used by a threat actor to enumerate and target accounts. This implies that the training data for LDAP classification may have multiple identical queries, often with conflicting labels. 

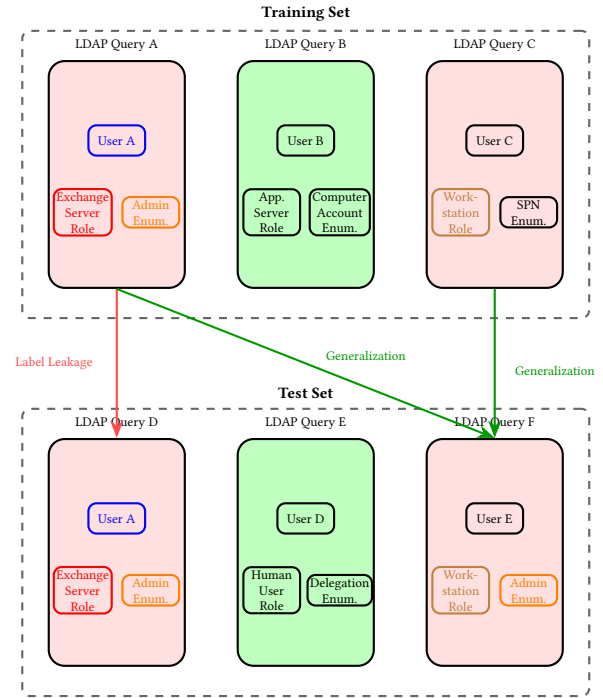
\begin{figure}[!h]
\centering
\begin{tikzpicture}[
    query/.style={rectangle, rounded corners=6pt, minimum width=1.8cm, minimum height=3cm, thick, draw=black},
    innerbox/.style={rectangle, rounded corners=3pt, draw=black, thick, align=center, font=\tiny, inner sep=1pt},
    pink/.style={fill=red!12},
    green/.style={fill=green!25},
    container/.style={rectangle, rounded corners=3pt, dashed, draw=black!60, thick},
    arrow/.style={-Stealth, thick}
]

\node[container, minimum width=7.5cm, minimum height=3.8cm] (train) at (0,0) {};
\node[above=0.01cm of train.north, font=\bfseries\scriptsize] {Training Set};

\node[query, pink] (qA) at (-2.5,0) {};
\node[above=0.01cm of qA.north, font=\tiny] {LDAP Query A};
\node[innerbox, blue, minimum width=0.75cm, minimum height=0.4cm] at (-2.5, 0.45) {User A};
\node[innerbox, red, minimum width=0.75cm, minimum height=0.4cm] at (-2.95, -0.5) {Exchange\\Server\\Role};
\node[innerbox, orange, minimum width=0.75cm, minimum height=0.4cm] at (-2.05, -0.5) {Admin\\Enum.};

\node[query, green] (qB) at (0,0) {};
\node[above=0.01cm of qB.north, font=\tiny] {LDAP Query B};
\node[innerbox, green, minimum width=0.75cm, minimum height=0.4cm] at (0, 0.45) {User B};
\node[innerbox, green, minimum width=0.75cm, minimum height=0.4cm] at (-0.45, -0.5) {App.\\Server\\Role};
\node[innerbox, green, minimum width=0.75cm, minimum height=0.4cm] at (0.45, -0.5) {Computer\\Account\\Enum.};

\node[query, pink] (qC) at (2.5,0) {};
\node[above=0.01cm of qC.north, font=\tiny] {LDAP Query C};
\node[innerbox, pink, minimum width=0.75cm, minimum height=0.4cm] at (2.5, 0.45) {User C};
\node[innerbox, brown, minimum width=0.75cm, minimum height=0.4cm] at (2.05, -0.5) {Work-\\station\\Role};
\node[innerbox, pink, minimum width=0.75cm, minimum height=0.4cm] at (2.95, -0.5) {SPN\\Enum.};

\node[container, minimum width=7.5cm, minimum height=3.8cm] (test) at (0,-5) {};
\node[above=0.01cm of test.north, font=\bfseries\scriptsize] {Test Set};

\node[query, pink] (qD) at (-2.5,-5) {};
\node[above=0.01cm of qD.north, font=\tiny] {LDAP Query D};
\node[innerbox, blue, minimum width=0.75cm, minimum height=0.4cm] at (-2.5, -4.55) {User A};
\node[innerbox, red, minimum width=0.75cm, minimum height=0.4cm] at (-2.95, -5.5) {Exchange\\Server\\Role};
\node[innerbox, orange, minimum width=0.75cm, minimum height=0.4cm] at (-2.05, -5.5) {Admin\\Enum.};

\node[query, green] (qE) at (0,-5) {};
\node[above=0.01cm of qE.north, font=\tiny] {LDAP Query E};
\node[innerbox, green, minimum width=0.75cm, minimum height=0.4cm] at (0, -4.55) {User D};
\node[innerbox, green, minimum width=0.75cm, minimum height=0.4cm] at (-0.45, -5.5) {Human\\User\\Role};
\node[innerbox, green, minimum width=0.75cm, minimum height=0.4cm] at (0.45, -5.5) {Delegation\\Enum.};

\node[query, pink] (qF) at (2.5,-5) {};
\node[above=0.01cm of qF.north, font=\tiny] {LDAP Query F};
\node[innerbox, pink, minimum width=0.75cm, minimum height=0.4cm] at (2.5, -4.55) {User E};
\node[innerbox, brown, minimum width=0.75cm, minimum height=0.4cm] at (2.05, -5.5) {Work-\\station\\Role};
\node[innerbox, orange, minimum width=0.75cm, minimum height=0.4cm] at (2.95, -5.5) {Admin\\Enum.};

\draw[arrow, red!70] (qA.south) -- node[left, xshift=-0.2cm, font=\tiny, text=red!70] {Label Leakage} (qD.north);
\draw[arrow, green!60!black] (qA.south) -- node[right, xshift=0.15cm, yshift=0.1cm, font=\tiny, text=green!60!black] {Generalization} (qF.north);
\draw[arrow, green!60!black] (qC.south) -- node[right, xshift=0.15cm, yshift=-0.1cm, font=\tiny, text=green!60!black] {Generalization} (qF.north);

\end{tikzpicture}
\caption{Label leakage example showing identical LDAP Query A and Query D in both training and test sets (red arrow), versus proper generalization where distinct queries with different characteristics appear across sets (green arrows).}
\label{fig:label_leakage}
\end{figure}

Having multiple identical queries presents a problem if left unaccounted for: label leakage. Figure \ref{fig:label_leakage} represents the problem of label leakage, where LDAP Query A is exactly the same as LDAP Query D and is present in both the training and test sets. In this example, the model is simply memorizing that when the Exchange Server User A executes this Admin Account Enumeration query, it is malicious. Since the model's goal is to generalize to query and user patterns to discover novel malicious LDAP reconnaissance queries, including the same user and query in both the test and training sets misrepresents its actual performance. An ideal scenario would resemble the green arrows: the model learns from LDAP Query A that Admin Enumeration is suspicious and from LDAP Query C that Workstation roles are suspicious, generalizing these patterns to predict Query F as malicious because it contains these 2 suspicious values.

\begin{figure}[!h]
\begin{tikzpicture}

\definecolor{crimson}{RGB}{220,20,60}
\definecolor{darkgray176}{RGB}{176,176,176}
\definecolor{green}{RGB}{0,128,0}
\definecolor{lightcoral}{RGB}{240,128,128}
\definecolor{lightgray204}{RGB}{204,204,204}
\definecolor{mediumpurple}{RGB}{147,112,219}
\definecolor{orange}{RGB}{255,165,0}
\definecolor{royalblue}{RGB}{65,105,225}

\begin{axis}[
width=\columnwidth,
height=0.4\columnwidth,
legend cell align={left},
legend style={
  fill opacity=0.8,
  draw opacity=1,
  text opacity=1,
  at={(0.5,-0.65)},
  anchor=north,
  draw=lightgray204,
  legend columns=3,
  font=\small,
  /tikz/every even column/.append style={column sep=0.3cm}
},
tick align=outside,
tick pos=left,
title={GroupShuffleSplit},
title style={font=\bfseries},
xlabel={Sample index},
xlabel style={font=\small},
xmin=0, xmax=100,
xtick style={color=black},
ymin=-0.35, ymax=1.35,
ytick style={color=black},
ytick={1,0},
yticklabels={Split,Group},
yticklabel style={font=\small},
enlargelimits=false,
clip=false
]
\draw[draw=white,fill=orange] (axis cs:0,0.85) rectangle (axis cs:20,1.15);
\addlegendimage{area legend,fill=orange}
\addlegendentry{Test set}

\draw[draw=white,fill=lightcoral] (axis cs:20,0.85) rectangle (axis cs:40,1.15);
\addlegendimage{area legend,fill=lightcoral}
\addlegendentry{Validation set}

\draw[draw=white,fill=royalblue] (axis cs:40,0.85) rectangle (axis cs:100,1.15);
\addlegendimage{area legend,fill=royalblue}
\addlegendentry{Training set}

\draw[draw=white,fill=blue] (axis cs:0,-0.15) rectangle (axis cs:20,0.15);
\draw[draw=white,fill=orange] (axis cs:20,-0.15) rectangle (axis cs:40,0.15);
\draw[draw=white,fill=green] (axis cs:40,-0.15) rectangle (axis cs:60,0.15);
\draw[draw=white,fill=crimson] (axis cs:60,-0.15) rectangle (axis cs:80,0.15);
\draw[draw=white,fill=mediumpurple] (axis cs:80,-0.15) rectangle (axis cs:100,0.15);
\end{axis}

\end{tikzpicture}
\caption{Grouped Splitting: A Process that splits the dataset into 5 folds, where no user is present in multiple folds. The first 3 folds are used for training, the 4th for validation, and the 5th for testing.}
\label{fig:groupshufflesplit}
\end{figure}
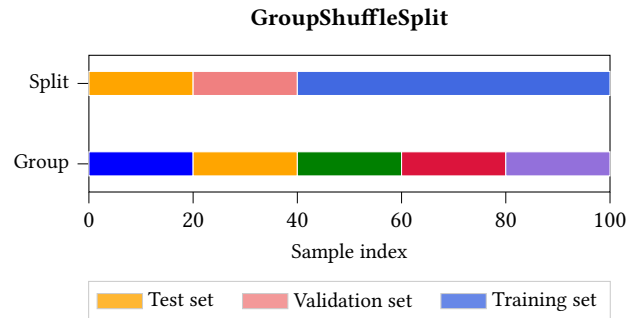

To prevent label leakage and ensure the model isn't simply memorizing source users or known patterns, a grouped splitting approach was used \cite{arlot2010survey, scikit-learn}. Figure \ref{fig:groupshufflesplit} displays the grouped splitting method taken to ensure no overlap in groups. In this case, the group is source users. Training and testing sets are split on different subsets of indices (or folds) of the sample set, where a user (the group in question) is not present in multiple splits. This encourages the model to learn to predict malicious LDAP query patterns instead of repeating memorized (known) patterns \cite{arlot2010survey, scikit-learn}. The grouped splitting methodology is a constraint that forces the model to generalize across different users. 

\subsubsection{Classification}
For downstream supervised classification, any classifier may be chosen. For this iteration, gradient-boosted decision trees were used, employing the XGBoost library \cite{10.1214/aos/1013203451, chen2016xgboost} due to their strong performance on the holdout set and their ability to handle class imbalance well relative to alternatives. 

Hyperparameter tuning was performed against the validation set using Bayesian optimization via hyperopt \cite{hyperopt,hyperopt_inproceedings}.

\subsubsection{Prediction Thresholds}
Prediction thresholds were chosen to control for the average number of daily false positives seen per customer. These thresholds were chosen based on the False Positive Rate (FPR) performance on the holdout test set, with the True Positive Rate (TPR) taken into account. 

These thresholds were designed around the goal of minimizing false 
positive alerts to customers so they would not exhibit alert fatigue, 
where too many false positive alerts diminish customer confidence in a 
detection. This is especially important when onboarding a completely new 
detection capability. Accordingly, all thresholds constrain to fewer than 
1 average false positive per day per customer.

A secondary benchmark of a minimum 30\% TPR was also established. 
Since weak labels are imperfect, the TPR benchmark is intentionally 
set lower than what would be expected of a fully supervised classifier 
to accommodate for that label noise. TPR was nonetheless included as a 
criterion because it demonstrates the model's ability to cover a higher 
volume of malicious LDAP queries. A higher TPR means more LDAP reconnaissance is caught.
\\

Threshold Constraints:
\begin{itemize}
    \item \textbf{Low}: Maximizes TPR while constraining average false positives to less than 1 per day per customer. 
    
    \item \textbf{Medium}: Maximizes TPR while constraining average false positives to less than 0.5 per day per customer. 

    \item \textbf{High}: Minimizes FPR while constraining TPR to at least 30\%.
\end{itemize}

Together, these thresholds allow for different alerting and severity 
options for LDAP reconnaissance detections, giving analysts flexibility 
to tune detection sensitivity to their operational environment.

\subsubsection{Predictions}

\begin{figure}[!h]
    \centering
    \includegraphics[width=1\linewidth]{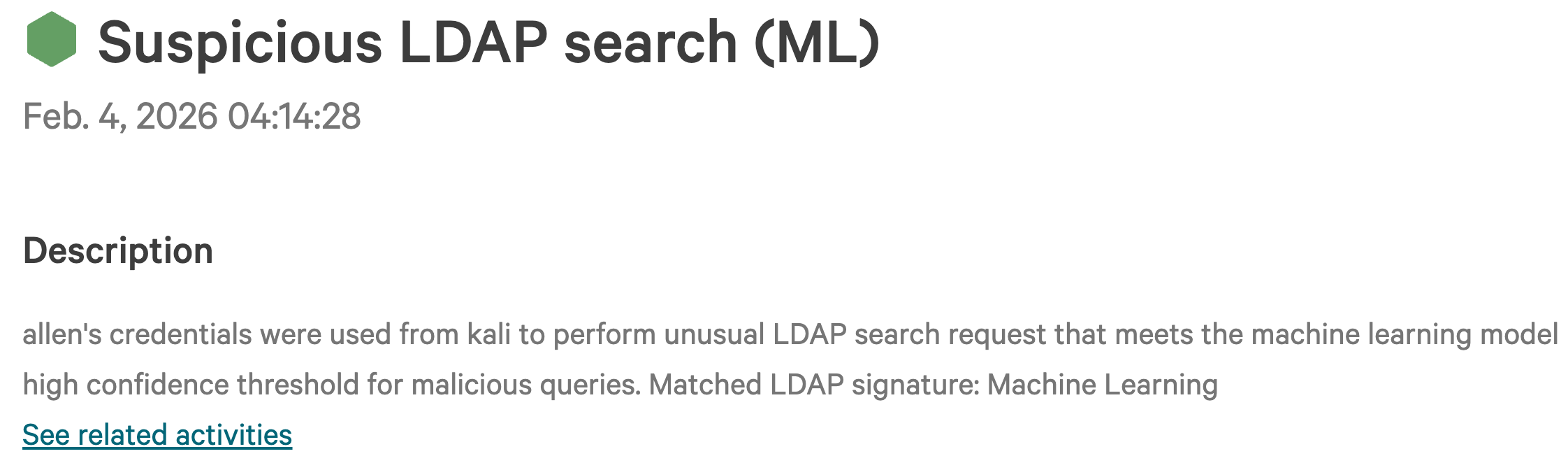}
    \caption{Sample LDAP ML Detection. This shows a sample Suspicious LDAP detection in the Falcon Identity Protection UI. The user can investigate this by analyzing the LDAP request, looking at correlated source account and endpoint security events, reviewing logs, and assessing the criticality of the target endpoints.}
    \label{fig:ldap_pred}
\end{figure}

Once the model is trained and validated, the classifier can now predict maliciousness for new LDAP queries across customers. Figure \ref{fig:ldap_pred} shows an example Suspicious LDAP search (ML) prediction in the CrowdStrike Falcon UI \cite{drew2025machine}. 

While the LDAP classifier model framework was the ultimate objective, it required developing additional infrastructure to support deployment and Machine Learning Operations (MLOps). Waiting for engineering teams to develop this infrastructure capability would leave customers exposed to LDAP reconnaissance in the meantime.

\section{Framework 2: ML-Based LDAP Signature Mining}
\label{sec:framework_2}

Given this temporary coverage gap, a new framework was developed to distill the results into a form compatible with standard infrastructure. This second framework, involving signature mining, could be immediately deployed within the current system while the new architectural requirements were being developed concurrently. This gave customers immediate protection against sophisticated LDAP attacks.
The strategy is to mine for new malicious signatures and plug them directly into the existing pipeline. 
Accordingly, a multi-step approach to data mining malicious LDAP signatures was developed.

\begin{enumerate}
    \item \textbf{Weak Labeling:} The 2nd framework uses the same weak labels used in the 1st framework. LDAP queries are joined with endpoint detections in close proximity for the same source user in order to produce a binary label. 
    
    \item \textbf{5 Downstream Classification Models:} Similar to the 1st framework, downstream classification is used to fit on weak labels and predict maliciousness. Signature mining, however, used 5 XGBoost classification models, each with different hyperparameters and thresholds, trained on different sets of data using 5-fold grouped cross-validation (CV), ensuring all of the sample data was mined for malicious signatures by concatenating the test fold predictions from each CV iteration. 
    
    \item \textbf{Signature Exclusion:} In addition to ML prediction, statistical hypothesis testing and family-wise error rate (FWER) exclusion was performed against malicious model predictions in order to validate signatures with high-confidence likelihood of maliciousness. 

\end{enumerate}

\subsection{Weak Labeling}
The weak labeling followed the exact same process as the first framework.

\subsection{5 Downstream Classification Models}

Downstream classification for signature mining involves a signature-hashing method, grouped cross-validation to prevent label leakage and encourage novel signature detection, and downstream classification using 5 XGBoost models trained on different partitions of the input data.

\subsubsection{Signature Hash Generation}
Different LDAP fields were concatenated into a single string and hashed in order to generate signature hashes. This hashing process was repeated with 2 different sets of fields included in the hash. This hashing assigned a unique signature to each LDAP query.

\textbf{Set 1 (Custom Hash)}: Created a signature based on the following LDAP search fields as well as the source entity role: \textsc{LdapSearchFilterShape}, \textsc{LdapSearchAttributes}, \textsc{LdapSearchScope}, \textsc{LdapSearchSizeLimit}, \textsc{Source Entity Roles}, \textsc{ActiveDirectoryAuthenticationMethod}, and \textsc{LdapSecurityType}.

\textbf{Set 2 (LdapSearchQueryToken)}: Created a signature based on the following LDAP search fields, ignoring user fields: \textsc{LdapSearchFilterShape}, \textsc{LdapSearchAttributes}, \textsc{LdapSearchScope}, and \textsc{LdapSearchSizeLimit}.

Now, each of the $\approx$1.3 million LDAP queries has 2 hashes assigned to it. 

\subsubsection{Grouped Cross-Validation by Signature Hash}

Signature mining splits the data in a similar fashion as the classifier in order to avoid label leakage. The classifier framework uses grouped splitting to ensure that the same user is not present in the training, validation, and test sets. The signature mining framework is concerned with discovering novel malicious hashes based on patterns it finds in other hashes. We don't want the LDAP signature hash to appear in both the train and test set. Therefore, it splits the datasets into 5 folds, with no LDAP query hash appearing in more than 1 fold. This is slightly different from the classifier framework, which only concerns itself with the same user not appearing in multiple folds. Recalling Figure \ref{fig:label_leakage}, this means that there won't be identical LDAP queries in the train and test set (query A and D). 

\begin{figure}[!h]

\begin{tikzpicture}

\definecolor{crimson}{RGB}{220,20,60}
\definecolor{darkgray176}{RGB}{176,176,176}
\definecolor{green}{RGB}{0,128,0}
\definecolor{lightcoral}{RGB}{240,128,128}
\definecolor{lightgray204}{RGB}{204,204,204}
\definecolor{mediumpurple}{RGB}{147,112,219}
\definecolor{orange}{RGB}{255,165,0}
\definecolor{royalblue}{RGB}{65,105,225}

\begin{axis}[
width=\columnwidth,
height=0.4\columnwidth,
legend cell align={left},
legend style={
  fill opacity=0.8,
  draw opacity=1,
  text opacity=1,
  at={(0.5,-0.65)},
  anchor=north,
  draw=lightgray204,
  legend columns=3,
  font=\small,
  /tikz/every even column/.append style={column sep=0.3cm}
},
tick align=outside,
tick pos=left,
title={GroupKFold},
x grid style={darkgray176},
xlabel={Sample index},
xmin=0, xmax=100,
xtick style={color=black},
y grid style={darkgray176},
ylabel={CV iteration},
ymin=-0.69, ymax=5.69,
ytick style={color=black},
ytick={0,1,2,3,4,5},
yticklabels={group,0,1,2,3,4}
]
\draw[draw=white,fill=orange] (axis cs:0,0.85) rectangle (axis cs:20,1.15);
\addlegendimage{area legend,fill=orange}
\addlegendentry{Test set}

\draw[draw=white,fill=lightcoral] (axis cs:20,0.85) rectangle (axis cs:40,1.15);
\addlegendimage{area legend,fill=lightcoral}
\addlegendentry{Validation set}

\draw[draw=white,fill=royalblue] (axis cs:40,0.85) rectangle (axis cs:100,1.15);
\addlegendimage{area legend,fill=royalblue}
\addlegendentry{Training set}

\draw[draw=white,fill=blue] (axis cs:0,-0.4) rectangle (axis cs:20,0.4);
\draw[draw=white,fill=orange] (axis cs:20,-0.4) rectangle (axis cs:40,0.4);
\draw[draw=white,fill=green] (axis cs:40,-0.4) rectangle (axis cs:60,0.4);
\draw[draw=white,fill=crimson] (axis cs:60,-0.4) rectangle (axis cs:80,0.4);
\draw[draw=white,fill=mediumpurple] (axis cs:80,-0.4) rectangle (axis cs:100,0.4);
\draw[draw=white,fill=royalblue] (axis cs:0,0.6) rectangle (axis cs:60,1.4);
\draw[draw=white,fill=orange] (axis cs:60,0.6) rectangle (axis cs:80,1.4);
\draw[draw=white,fill=lightcoral] (axis cs:80,0.6) rectangle (axis cs:100,1.4);
\draw[draw=white,fill=royalblue] (axis cs:0,1.6) rectangle (axis cs:40,2.4);
\draw[draw=white,fill=orange] (axis cs:40,1.6) rectangle (axis cs:60,2.4);
\draw[draw=white,fill=lightcoral] (axis cs:60,1.6) rectangle (axis cs:80,2.4);
\draw[draw=white,fill=royalblue] (axis cs:80,1.6) rectangle (axis cs:100,2.4);
\draw[draw=white,fill=royalblue] (axis cs:0,2.6) rectangle (axis cs:20,3.4);
\draw[draw=white,fill=orange] (axis cs:20,2.6) rectangle (axis cs:40,3.4);
\draw[draw=white,fill=lightcoral] (axis cs:40,2.6) rectangle (axis cs:60,3.4);
\draw[draw=white,fill=royalblue] (axis cs:60,2.6) rectangle (axis cs:100,3.4);
\draw[draw=white,fill=orange] (axis cs:0,3.6) rectangle (axis cs:40,4.4);
\draw[draw=white,fill=lightcoral] (axis cs:20,3.6) rectangle (axis cs:40,4.4);
\draw[draw=white,fill=royalblue] (axis cs:40,3.6) rectangle (axis cs:100,4.4);
\draw[draw=white,fill=lightcoral] (axis cs:0,4.6) rectangle (axis cs:20,5.4);
\draw[draw=white,fill=royalblue] (axis cs:20,4.6) rectangle (axis cs:80,5.4);
\draw[draw=white,fill=orange] (axis cs:80,4.6) rectangle (axis cs:100,5.4);
\end{axis}

\end{tikzpicture}
\caption{GroupKFold Cross-Validation Modeling: For each iteration, 3 folds are used to train a model, 1 fold for validation, and 1 fold for testing. The datasets shift folds every CV iteration in order to have the test set cover the entire data sample after 5 iterations.}
\label{fig:groupkfold_custom}
\end{figure}
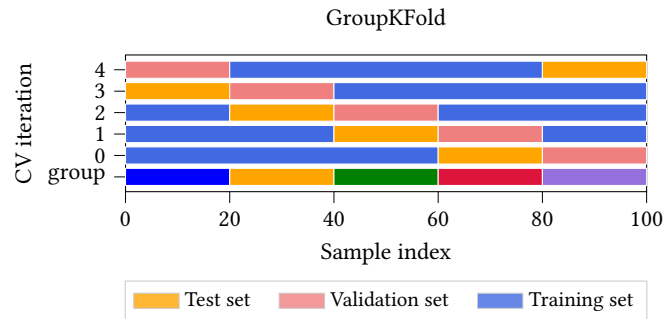

A new problem also arises in signature mining, where predicting malicious signatures on the test set is only going to give predictions for 20\% of the data. Since we want to mine the entire dataset for malicious signatures, we take an iterative cross-validation approach known as GroupKFold Cross-Validation (CV). To avoid fitting and predicting on the same data, we use 5 different CV iterations so that the holdout test set shifts and covers a different 20\% of the data with each iteration. This allows prediction of the whole sample set after 5 iterations. Figure \ref{fig:groupkfold_custom} shows how, during each CV iteration, 3 folds are used to train the model, 1 fold for validation, and 1 fold for testing. For each iteration, a separate XGBoost model is fit; thereby, preventing any label leakage during training, hyperparameter, and threshold tuning. After all 5 models are fit to the training sets and applied to the test sets, the test-set predictions are concatenated. This 5-model approach allows us to mine the entire dataset without any leakage. The grouped approach ensures that it can generalize to unseen signatures rather than memorizing known malicious signatures.

\subsubsection{Hyperparameter Tuning}
If the same hyperparameters learned in 1 iteration were used for the remaining 5 iterations, this would introduce label leakage, since the hyperparameters were learned on data intended to be held out in that iteration. As a solution, the validation fold is used for hyperparameter tuning at each iteration, yielding 5 distinct hyperparameter sets based on the 5 validation sets. As with the modeling process, Hyperopt was used on the validation set to select the optimal hyperparameter configuration for each model \cite{hyperopt}. 5 different models were fit using 5 different sets of hyperparameters, ensuring that no label leakage was incorporated into the hyperparameters. 

\subsubsection{Threshold Tuning}
\textit{}
\newline
A single threshold decision boundary was set for each model in this process. This decision boundary was similar to the "Low" threshold in the modeling threshold-tuning process, except that it was selected using the validation set rather than the test set. On the validation set, the model selected a threshold that maximized TPR while ensuring an average of fewer than 1 FP per day per customer. This ensured a threshold that aligned with expectations but was less stringent than previous thresholds. Given additional downstream signature-exclusion criteria, we opted to cast a wider initial net for maliciousness prediction. 

\begin{figure}[!h]
    \centering
    \begin{tikzpicture}
    \begin{axis}[
        width=0.8\columnwidth,
        height=0.8\columnwidth,
        title={Receiver Operating Characteristic (ROC) Curve},
        title style={font=\bfseries},
        xlabel={False Positive Rate},
        ylabel={True Positive Rate},
        xmin=0, xmax=1,
        ymin=0, ymax=1,
        grid=major,
        grid style={dashed,gray!30},
        legend pos=south east,
        xlabel style={font=\small},
        ylabel style={font=\small},
        tick label style={font=\small}
    ]

    \addplot[
        color=blue,
        line width=1.5pt,
        smooth
    ] coordinates {
        (0.00, 0.00)
        (0.01, 0.22)
        (0.02, 0.32)
        (0.03, 0.40)
        (0.05, 0.47)
        (0.07, 0.54)
        (0.08, 0.57)
        (0.10, 0.61)
        (0.12, 0.64)
        (0.15, 0.67)
        (0.20, 0.72)
        (0.25, 0.76)
        (0.30, 0.79)
        (0.40, 0.84)
        (0.50, 0.88)
        (0.60, 0.92)
        (0.70, 0.95)
        (0.80, 0.97)
        (0.90, 0.99)
        (1.00, 1.00)
    };

    \addplot[
        only marks,
        mark=*,
        mark size=3pt,
        color=red
    ] coordinates {(0.08, 0.57)};

    \addplot[
        dashed,
        gray,
        line width=0.8pt
    ] coordinates {(0,0) (1,1)};
    
    \end{axis}
    \end{tikzpicture}
    \caption{Example LDAP Model ROC Curve for Validation Set. This shows the Receiver Operating Characteristic (ROC) curve for a validation set during one of the iterations. This threshold (red point) was chosen to maximize the TPR at a constrained minimum acceptable FPR. This validation set showed a .57 TPR and a .08 FPR.}
    \label{fig:roc_curve}
\end{figure}
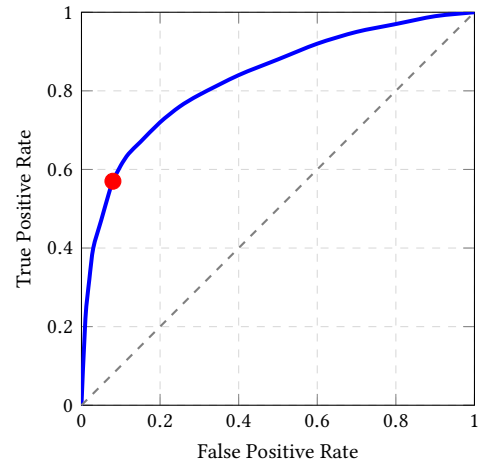

Figure \ref{fig:roc_curve} shows an example Receiver Operating Characteristic (ROC) Curve with a threshold selected against the validation set. This constrained the average FPs to 1 per day per customer, while maximizing TPR, resulting in 57\% validation TPR.  

\subsubsection{Predictions}
After the 5 models predict malicious LDAP signatures on the test sets, these predictions and their predicted probabilities are combined for data mining. 

\subsection{Signature Exclusion via Hypothesis Testing}

 The main concern regarding immediate deployment is the proliferation of false positives in customer environments. This would cause analyst fatigue and distrust in this new detection capability. For the signature mining framework, ensuring minimal false positives involved a 2-step process: binomial hypothesis testing followed by Family-Wise Error Rate (FWER) Exclusion. 

\subsubsection{One-Tailed Binomial Hypothesis Test} 

The statistical framework for evaluating signature efficacy relies on a one-tailed binomial hypothesis test to determine whether a signature exhibits bias toward malicious LDAP queries.
$H_0: p \leq p_0$
defines the null hypothesis, which assumes the signature is not biased toward malicious queries,
where $p$ represents the true proportion of malicious queries flagged by the signature, and $p_0$ represents the baseline proportion of malicious queries in the sample population.
$H_1: p > p_0$
establishes the alternative hypothesis, indicating the signature demonstrates bias toward malicious queries.
For a binomial test with $n$ total queries and $x$ observed malicious queries flagged by the signature,
$X \sim \text{Binomial}(n, p_0)$
is the test statistic of interest.
The corresponding p-value$ = P(X \geq x | H_0) = \sum_{k=x}^{n} \binom{n}{k} p_0^k (1-p_0)^{n-k}$.

The decision criterion is established such that if $\text{p-value} < 0.05$, we reject $H_0$ and conclude there is less than a 5\% chance of incorrectly labeling the signature as biased when it is not.
$\text{Power} = P(\text{reject } H_0 | H_1 \text{ is true}) = 1 - \beta$
expresses the statistical power of the test, which represents the probability of correctly detecting bias when it truly exists,
where $\beta$ is the Type II error rate. 
At the conclusion of the hypothesis tests, only signatures with a $\text{p-value} < 0.05$ and $\text{power} > 0.80$ are filtered through.

\subsubsection{Family-Wise Error Rate (FWER) Exclusion}

The second step for excluding signatures employs Family-Wise Error Rate (FWER) Exclusion, which uses a strict statistical framework to ensure that only signatures with acceptable false-positive rates are retained \cite{westfall1993resampling}. In order to ensure that the average false positive rate of a given signature is less than 1 per day per customer, FWER was used to exclude signatures with $\text{false positive rates} \geq 1$ per day for \textbf{any} relevant customers. Even if a single customer has more than 1 FP/day, this signature will be filtered out. 

The statistical foundation begins with establishing the null and alternative hypotheses for each signature-customer combination. 

$H_0: \mu_{FP} \geq 1$
defines the null hypothesis, where $\mu_{FP}$ represents the mean daily false positive rate for a given signature-customer pair. The corresponding alternative hypothesis is $H_1: \mu_{FP} < 1$.

To control the family-wise error rate across multiple simultaneous tests, we apply the Bonferroni correction $\alpha_{corrected} = \frac{\alpha}{m}$\cite{bonferroni1936teoria, holm1979simple} for $m$ customers being tested for a particular signature, where $\alpha$ is the desired family-wise error rate (typically 0.05).

For each customer, we transform the observed daily false positive counts by subtracting the threshold value plus a small epsilon to handle zero differences. The test statistic calculation is
$d_i = x_i - (\mu + \epsilon)$, where $x_i$ represents the observed daily false positive count for day $i$, $\mu = 1$ is our threshold, and $\epsilon = 10^{-10}$ prevents zero differences.

The Wilcoxon signed-rank test statistic $W = \sum_{i: d_i > 0} R_i$ is computed as the sum of ranks corresponding to positive differences, where $R_i$ is the rank of $|d_i|$ among all non-zero absolute differences.

The significance determination for each customer follows the criterion in Equation \eqref{eq:significance_test}:

\begin{equation}
\text{Significant}_C = \begin{cases}
\text{True} & \text{if } p_C < \alpha_{corrected} \\
\text{False} & \text{otherwise}
\end{cases}
\label{eq:significance_test}
\end{equation}

A signature is retained only if all customers show significant results, as defined in Equation \eqref{eq:signature_retention}:

\begin{equation}
\text{Retain Signature} = \bigwedge_{j=1}^{m} \text{Significant}_{C_j}
\label{eq:signature_retention}
\end{equation}

The proportion of significant customers for a signature is calculated using Equation \eqref{eq:sig_proportion}:

\begin{equation}
\text{Significance Proportion} = \frac{\sum_{j=1}^{m} \mathbf{1}(\text{Significant}_{C_j})}{m}
\label{eq:sig_proportion}
\end{equation}

where $\mathbf{1}(\cdot)$ is the indicator function.
Finally, the minimum alpha value that would result in signature retention is computed as $\alpha_{min} = m \cdot \max_{j=1}^{m} p_{C_j}$.

This value represents the family-wise error rate at which the signature would just barely pass the exclusion criteria, providing insight into the signature's statistical robustness. 
For a given signature, if, for any customer, we fail to reject the null hypothesis, then that signature is excluded.

\paragraph{Sensitivity Analysis.}
To confirm the robustness of the FWER filtering criteria, we conducted 
a sensitivity analysis across both the alpha and power thresholds.

Table~\ref{tab:fwer_alpha_sensitivity} shows the alpha sensitivity 
results. The retained signature set is identical across all tested 
alpha levels, with zero alpha-sensitive signatures. All 54 signatures 
have p-values $< 0.001$, confirming that no signature is a borderline 
statistical case and that the choice of $\alpha = 0.05$ has no 
material effect on the output.

\begin{table}[!h]
\centering
\caption{Alpha sensitivity analysis. All 54 signatures are retained 
at every tested alpha level, confirming no signature is a borderline 
statistical case. The production threshold ($\alpha = 0.050$) is 
shown in bold.}
\label{tab:fwer_alpha_sensitivity}
\small
\begin{tabular}{cc}
\toprule
\textbf{Alpha} & \textbf{Retained} \\
\midrule
0.001 & 54 \\
0.010 & 54 \\
\textbf{0.050} & \textbf{54} \\
0.100 & 54 \\
0.200 & 54 \\
\bottomrule
\end{tabular}
\end{table}

Table~\ref{tab:fwer_power_sensitivity} shows the power sensitivity 
results. The power threshold reveals a clean precision-coverage 
tradeoff: stricter thresholds yield monotonically higher precision 
at the cost of fewer retained signatures. Power-sensitive signatures 
are excluded by effect size rather than sample size. Signatures 
dropped at higher power thresholds have lower average malicious query 
rates despite similar query counts, confirming the threshold functions 
as a genuine effect size filter. The standard power $= 0.80$ 
convention represents the appropriate 
balance for this deployment context \cite{cohen1988statistical}. The pipeline rejects 99.3\% of 
all candidate signatures overall (7,705 to 54), confirming that the 
filtering criteria are stringent rather than permissive.

\begin{table}[!h]
\centering
\caption{Power sensitivity analysis. Stricter power thresholds reveal 
a clean precision-coverage tradeoff. Power-sensitive signatures are 
excluded by effect size rather than sample size. The production 
threshold (power $= 0.80$) is shown in bold.}
\label{tab:fwer_power_sensitivity}
\small
\begin{tabular}{ccc}
\toprule
\textbf{Power} & \textbf{Retained} & \textbf{Precision \%} \\
\midrule
0.60 & 54 & 84 \\
0.70 & 54 & 84 \\
\textbf{0.80} & \textbf{54} & \textbf{84} \\
0.90 & 35 & 92 \\
0.95 & 22 & 95 \\
\bottomrule
\end{tabular}
\end{table}

\subsubsection{Predictions}
After signature exclusion, all remaining signatures are flagged as malicious. After running this process end-to-end for the custom hash, 44 high-confidence signatures were generated. After running it end-to-end for the LdapSearchQueryToken hash, 10 high-confidence signatures were generated. This resulted in a total of 54 high-confidence ML-mined signatures. 

\section{Evaluation and Results}

Frameworks 1 and 2 were evaluated slightly differently: TPR and FPR were the main criteria for the classifier, whereas precision and FPR were the main criteria for mined signatures.

\subsection{Framework 1: Classifier Evaluation}

The classifier evaluation considers the following metrics in measuring model performance: FPR, average daily FPs per customer, AUC ROC, flag rate, and precision. The primary metrics of interest are TPR and FP/day/customer, as these were identified by Identity Protection (IDP) experts as Key Performance Indicators (KPIs) for the business. IDP preferred a classifier $\geq$30\% TPR in order to ensure that the model captures enough signal. This is lower than the standards of traditional classifiers due to weak labeling. They also preferred an FPR below a rate that corresponds to 1 false positive per day per customer, which would solve any concerns around false positives creating alert fatigue for analysts.

\subsubsection{Evaluation Against Historical Corpus}

A historical corpus was collected that included $\approx$6 billion LDAP queries, 90 days, and 367 customers. The ML classifier only predicted on uncommon signatures. After filtering for uncommon signatures, the corpus comprised a sample of $\approx$1.3 million queries, 8,344 of which were malicious according to the weak labels. The following evaluations are measured against the holdout test dataset, which consisted of 259,012 LDAP queries, 2,540 labeled malicious.

In Table \ref{tab:ml_performance_metrics}, different performance metrics were evaluated at pre-defined decision thresholds and compared to the benchmark. In addition, we offered alternative model comparisons, Random Forest (RF) and Logistic Regression (LR). Each threshold exceeded benchmark goals for the primary XGBoost model, outperforming both TPR and FPR expectations and outperforming alternative ML models. The low threshold achieved a TPR of 65\%, exceeding the medium threshold (52\%) by 13\%, more than doubling the 30\% benchmark requirement. The high threshold exhibited the most conservative false positive rate at 0.17 FP/day/customer (corresponding to 1\% FPR), representing an 83\% reduction relative to the 1.0 benchmark.

\begin{table*}[!t]
\caption{Classifier Performance Metrics: Evaluates classifier performance at different thresholds of low, medium, and high, with XGBoost demonstrating exceptional performance compared to the benchmark Identity Expectation and alternative ML model ablations across all thresholds.}
\centering
\small
\begin{tabular}{lcllllllccc}
\toprule
\textbf{Metric} & \textbf{Identity} & RF & RF & RF & LR & LR & LR & \textbf{XGB} & \textbf{XGB} & \textbf{XGB}\\
& \textbf{Expectation} & High & Medium & Low & High & Medium & Low & \textbf{High} & \textbf{Medium} & \textbf{Low} \\
\midrule
TPR & 30\% & 30\% & 53\% & 58\% & 30\% & 42\% & 56\% & 30\% & 52\% & 65\% \\
FP/day/customer & 1 & 0.19 & 0.50 & 1.0 & 0.24 & 0.50 & 1.0 & 0.17 & 0.48 & 0.99 \\
FPR & NA & 1.2\% & 3.3\% & 6.6\% & 1.6\% & 3.3\% & 6.6\% & 1\% & 3.2\% & 6.6\% \\
AUC ROC & NA & 0.85 & 0.85 & 0.85 & 0.81 & 0.81 & 0.81 & 0.85 & 0.85 & 0.85 \\
Flag Rate & NA & 1.5\% & 3.9\% & 7.6\% & 1.9\% & 3.8\% & 7.5\% & 1.4\% & 3.8\% & 7.6\% \\
Precision & NA & 19\% & 14\% & 8\% & 16\% & 11\% & 7.7\% & 21\% & 14\% & 9\% \\
\bottomrule
\end{tabular}
\label{tab:ml_performance_metrics}
\end{table*}

Table \ref{tab:confusion_results} shows the confusion matrices for each decision threshold on the test data. 

\begin{table}[!h]
\centering
\caption{Classifier performance at all three production thresholds of interest. }
\label{tab:confusion_results}
\begin{tabular}{@{}lcrcrcr@{}}
\toprule
Threshold & \multicolumn{2}{c}{High} & \multicolumn{2}{c}{Medium} & \multicolumn{2}{c}{Low} \\ \cmidrule(lr){2-3} \cmidrule(lr){4-5}  \cmidrule(l){6-7} 
True \textbackslash Pred. &
  Ben. &
  \multicolumn{1}{c}{Mal.} &
  Ben. &
  \multicolumn{1}{c}{Mal.} &
  Ben. &
  \multicolumn{1}{c}{Mal.} \\ \midrule
Benign &
  \multicolumn{1}{r}{253,539} &
  2,933 &
  \multicolumn{1}{r}{248,324} &
  8,148 &
  \multicolumn{1}{r}{239,623} &
  16,849 \\
Malicious &
  \multicolumn{1}{r}{1,779} &
  761 &
  \multicolumn{1}{r}{1,219} &
  1,321 &
  \multicolumn{1}{r}{884} &
  1,656 \\ \bottomrule
\end{tabular}
\end{table}

\subsection{Framework 2: Signature Mining Evaluation}

The signature mining evaluation process also used the same metrics but placed greater emphasis on minimizing FPR. The goal was to mine high-precision, low-FPR signatures for immediate deployment with minimal customer complaints, while the more general classifier awaited official deployment.

\subsubsection{Historical Evaluation}

The same historical corpus was used for signature mining in both efforts, with the LdapSearchQueryToken hash and the custom hash. The dataset contained 48,470 unique LdapSearchQueryToken hashes and 68,130 custom hashes. 

A precursor to signature mining was the prediction on 5 CV iterations using 5 downstream XGBoost classifiers. Table \ref{tab:cv_results} shows the performance of these 5 models across all 5 iterations, reporting test-set performance for each iteration. 

\begin{table}[!h]
\caption{Cross-validation results showing test set performance metrics for the 5 XGBoost models in framework 2 across 5 iterations}
\centering
\begin{tabular}{|c|c|c|c|c|}
\hline
Iteration & TPR & FP/day/customer & FPR & AUC ROC \\
\hline
0 & 0.44 & 0.91 & 0.07 & 0.78 \\
1 & 0.58 & 1.06 & 0.08 & 0.81 \\
2 & 0.41 & 0.82 & 0.06 & 0.78 \\
3 & 0.50 & 1.19 & 0.09 & 0.79 \\
4 & 0.54 & 1.14 & 0.09 & 0.79 \\
\hline
\end{tabular}

\label{tab:cv_results}
\end{table}

Table \ref{tab:test_range_table} summarizes these ranges, which generally adhere to and exceed expectations. 

\begin{table}[!h]
\centering
\caption{Test-Fold Evaluation. Shows the range of performance metrics across the 5 test folds for each of the 5 models}
\begin{tabular}{|l|c|c|}
\hline
\textbf{Metric} & \textbf{Identity\ Expectation} & \textbf{Test Range} \\
\hline
TPR & 30\% & 41.2\% - 57.6\% \\
\hline
FP/day/customer & 1 & 0.82 - 1.19 \\
\hline
FPR & NA & 6.3\% - 9.0\% \\
\hline
AUC ROC & NA & 0.78 - 0.81 \\
\hline
\end{tabular}
\label{tab:test_range_table}
\end{table}

This, however, precedes statistical signature exclusion via hypothesis testing. After FWER adjustment, we can expect that no signatures have a daily false-positive rate greater than 1 for any given customer. Post-signature exclusion, the combined precision for remaining signatures was calculated on the test set. We achieve a \textit{test precision of 84\% across the 54 signatures mined}. This includes 10 LdapSearchQueryToken hash signatures and 44 custom hash signatures.

\subsubsection{Live Evaluation}

In live evaluation, CrowdStrike's Falcon Complete Managed Detection and Response (MDR) analysts triaged each 
triggered detection and labeled it TP or FP using standard 
security investigation procedures. Because analysts only review 
flagged predictions, TPR cannot be computed on live data; 
precision is therefore the primary production metric.

Table~\ref{tab:live_falcon_ldap_efficacy} compares holdout 
precision (evaluated against heuristic weak labels) with live 
human-labeled precision from MDR field deployment. 
The holdout precision of 84.07\% and live precision of 81.48\% 
have overlapping 95\% confidence intervals, demonstrating that 
holdout evaluation is a reliable proxy for production performance. 
All confidence intervals are 95\% Wilson score intervals, which 
provide reliable coverage at small sample sizes~\cite{Wilson01061927}.

These new ML-mined signatures achieved higher precision than 
approximately 73.5\% of identity detections live during the same 
period, indicating strong operational value relative to the 
broader detection portfolio.

\begin{table}[!h]
\centering
\small
\caption{Falcon Complete Efficacy for LDAP ML-Mined Signatures. Displays both 
heuristic-labeled efficacy against the holdout set and human-labeled efficacy 
against production data.}
\begin{tabular}{lcccc}
\toprule
\textbf{Metric} & 
\textbf{n} & 
\textbf{Customers} & 
\textbf{Precision} & 
\textbf{95\% CI} \\
\midrule
ML-Mined Signatures\\ (Holdout)   
    & 979 & 367 & 84.07\% & [81.6\%, 86.2\%] \\
ML-Mined Signatures\\ (Live Human-Labeled)  
    & 81  & 42  & 81.48\% & [71.7\%, 88.4\%] \\
\bottomrule
\end{tabular}
\label{tab:live_falcon_ldap_efficacy}
\end{table}

\section{Conclusion}

Both frameworks successfully address the limitations of traditional 
signature-based LDAP reconnaissance detection. The ML classifier exceeded 
performance expectations across all decision thresholds, achieving True Positive Rates 
between 30--65\% while maintaining False Positive Rates well below the 
1-per-day-per-customer benchmark, with an AUC ROC of 0.85 and the 
flexibility to tune sensitivity to operational needs. The signature mining 
framework proved equally effective in production, delivering 81.48\% field 
precision and outperforming 73.5\% of existing identity detections. Its 54 
high-confidence signatures filled an immediate coverage gap while the 
classifier awaited deployment, protecting customers from reconnaissance 
attempts that would have otherwise gone undetected.

Both frameworks also demonstrate a practical approach to weak supervision 
at scale: by correlating endpoint detections with LDAP queries, millions of 
training labels were generated without the bottleneck of manual labeling. 
The primary deployment challenge was the ML classifier requiring new MLOps 
infrastructure not yet available. This motivated Framework~2, which distills ML 
model outputs into signatures compatible with existing rules-based 
infrastructure. The key generalizable lesson is the dual-framework 
itself: \textit{train a full ML model, then mine deployable artifacts from 
it as a bridging solution}. This pattern applies to any domain 
where ML infrastructure is immature but rules-based deployment pipelines 
already exist. In cybersecurity, this decouples ML development timelines from customer protection. Finally, this methodology can be extended to other identity protocols where reconnaissance correlates with broader attack patterns, offering a template for adaptive ML-powered detections beyond known attack tool signatures.

\bibliographystyle{ACM-Reference-Format}
\balance
\bibliography{references,sample-base}

\appendix
\label{app:a}

\section{LDAP Tutorial} \label{sec:ldap_tutorial}

The dataset comprises LDAP queries and their metadata. Below is an example of a malicious LDAP query that searches for non-domain controller computers with Resource-Based Constrained Delegation (RBCD) configured, a configuration that attackers use to identify targets for privilege escalation attacks.
\begin{verbatim}
ldapsearch 
\end{verbatim}
\noindent{\texttt{-H ldap://dc01.example.com:389}} \\
\textcolor{authcolor}{\texttt{-Y GSSAPI}} \\
\textcolor{securitycolor}{\texttt{-O minssf=1,maxssf=1}} \\
\texttt{-b "DC=example,DC=com"} \\
\textcolor{scopecolor}{\texttt{-s sub}} \\
\textcolor{limitcolor}{\texttt{-z 1000}} \\
\textcolor{filtercolor}{\btt{"(\&(msds-allowedtoactonbehalfofotheridentity=*)}} \\
\textcolor{filtercolor}{\texttt{(!(primarygroupid=516)))"}} \\
\textcolor{attrcolor}{\texttt{msds-allowedtoactonbehalfofotheridentity sAMAccountName distinguishedName}}

This query establishes a connection to the domain controller \texttt{dc01.example.com} on the standard LDAP port 389. The user authenticates using \textcolor{authcolor}{Kerberos authentication} (\texttt{-Y GSSAPI}), which is captured in our dataset as the \textbf{ActiveDirectoryAuthenticationMethod} field.

The query specifies \textcolor{securitycolor}{SASL integrity protection} \\(\texttt{-O minssf=1,maxssf=1}), ensuring messages are integrity-protected. This configuration is recorded as the \textbf{LdapSecurityType} field with value \texttt{SASL\_INTEGRITY}, indicating the security method applied to the LDAP connection.

Starting from the base distinguished name \texttt{"DC=example,DC=com"}, the query searches with \textcolor{scopecolor}{\texttt{-s sub}} scope, meaning it traverses the entire directory subtree. This is captured as the \textbf{LdapSearchScope} field with value \texttt{WHOLE\_SUBTREE}, which matches the base object itself and all its subordinates.

The query limits results to \textcolor{limitcolor}{1000 entries} (\texttt{-z 1000}), recorded as the \textbf{LdapSearchSizeLimit} field. A value of 0 would indicate no limit.

The \textcolor{filtercolor}{search filter} is the query filter itself. The filter:\\ \texttt{(\&(msds-allowed\-to\-act\-on\-behalf\-of\-other\-identity=*)}\\\texttt{(!(primarygroupid=516)))} looks for objects with two conditions: they must have the \texttt{msds-allowedtoactonbehalfofotheridentity} attribute set (indicating RBCD is configured) AND NOT belong to the primary group 516 (Domain Controllers). The negation excludes domain controllers, likely targeting workstations and member servers (ideal privilege escalation targets).

This filter is normalized and stored as the \textbf{LdapSearchFilterShape} field: \btt{(\&(msds-allowedtoactonbehalfofotheridentity=*)(!(primarygroupid=\_)))}. In this normalized form, specific values are replaced with \texttt{\_} (representing concrete objects) while wildcards remain as \texttt{*} (representing multiple objects). This abstraction enables pattern matching across similar queries regardless of specific values.

The \textcolor{attrcolor}{requested attributes} (\btt{msds-allowedtoactonbehalfofotheridentity}, \texttt{sAMAccountName}, and \texttt{distinguishedName}) reveal exactly what information the user wants to extract. This would be the equivalent of specifying the fields you want returned in a SQL query. These are captured in the \textbf{LdapSearchAttributes} field as a distinct, sorted list. The combination of these specific attributes is highly indicative of delegation abuse reconnaissance.

Our system automatically assigns this query rules-based classifications through the \textbf{LdapSearchQueryClassification} field: \texttt{DELEGATION\_ENUMERATION} and \texttt{COMPUTER\_ACCOUNT\_ENUMERATION}. These tags surface specific behaviors that have potential to be used for reconnaissance.

Finally, the dataset includes \textbf{Source Entity Roles} for the account that executed the query. In this case, if the query came from a user account, it would be tagged as \texttt{HumanUserRole}. This computed field distinguishes between human users, service accounts (\texttt{ProgrammaticUserRole}), and other entity types, helping analysts understand whether the activity aligns with the account's expected behavior patterns.

Features were derived from the fields in Table \ref{tab:ldap-fields}. For \textcolor{filtercolor}{\btt{LdapSearchFilterShape}}, the text was tokenized and the Term Frequency-Inverse Document Frequency (TF-IDF) was calculated for each token and bigram. The text from \textcolor{attrcolor}{\btt{LdapSearchAttributes}} was also tokenized, calculating the TF-IDF of only the single-word tokens. \textcolor{scopecolor}{\btt{LdapSearchScope}} was numerically encoded based on the scope size; 0 representing a match for the base object itself, 1 representing a single level, and 2 representing the whole subtree. \textcolor{limitcolor}{\btt{LdapSearchSizeLimit}} kept the raw limit number. \btt{LdapSearchQueryClassification} features were just count vectors of the classifications, numerically encoded. \btt{LdapSearchQueryToken} wasn't used as a feature in the model; however, it was used later on in the signature mining application. \textcolor{securitycolor}{\btt{LdapSecurityType}} and \textcolor{authcolor}{\btt{Active\-Directory\-Authentication\-Method}} were one-hot encoded. \btt{Source Entity Roles} were numerically encoded with count vectors. 

\begin{table*}[!h]
    \centering
    \caption{LDAP Activity Field Definitions}
    \resizebox{\textwidth}{!}{
    \begin{tabular}{|p{4cm}|p{4cm}|p{5cm}|}
        \hline
        \textbf{Field} & \textbf{Example Value} & \textbf{Definition} \\
        \hline
        \textcolor{filtercolor}{\textbf{LdapSearchFilterShape}} & 
        \btt{(\&(msds-allowedtoactonbehalfofotheridentity=*)(!(primarygroupid=_)))} & 
        The LDAP search query filter associated with the activity. Specific objects are obfuscated, with "\_" representing specific objects and "*" representing multiple objects.  \\
        \hline
        \textcolor{attrcolor}{\textbf{LdapSearchAttributes}} & 
        \btt{msds-allowedtoactonbehalfofotheridentity sAMAccountName distinguishedName} & 
        Distinct, sorted list of projected LDAP attributes associated with the activity \\
        \hline
        \textcolor{scopecolor}{\textbf{LdapSearchScope}} & 
        \btt{WHOLE_SUBTREE} & 
        The LDAP search scope - defines search depth (base, one level, or subtree). In this example, the scope is the whole subtree, which matches the base object itself and any of its subordinates. \\
        \hline
        \textcolor{limitcolor}{\textbf{LdapSearchSizeLimit}} & 
        \btt{1000} & 
        The LDAP size-limit for query results. 0 implies no limit \\
        \hline
        \textbf{LdapSearchQueryClassification} & 
        \btt{DELEGATION_ENUMERATION\  COMPUTER_ACCOUNT_ENUMERATION} & 
        Rule-based classifications for the LDAP query. This classifies the query as an enumeration of computer accounts and an enumeration of accounts that have delegation capabilities. \\
        \hline
        \textbf{LdapSearchQueryToken} & 
        6ba7b810-9dad-11d1-80b4-00c04fd430c8 & 
        Hash of the LDAP query combining filter shape, attributes, scope, and size-limit \\
        \hline
        \textcolor{securitycolor}{\textbf{LdapSecurityType}} & 
        \btt{SASL_INTEGRITY} & 
        Method for securing LDAP messages.  \\
        \hline
        \textcolor{authcolor}{\textbf{Active\-Directory\-Authentication\-Method}} & 
        \btt{KERBEROS} & 
        The mechanism used for authenticating with an Active Directory domain. \\
        \hline
        \textbf{Source Entity Roles} & 
        HumanUserRole & 
        Computed field which assigns a notion of class to the user entity who executed the LDAP query. e.g. HumanUserRole, ProgrammaticUserRole, etc. \\
        \hline
    \end{tabular}
    }
    \label{tab:ldap-fields}
\end{table*}

\end{document}